# Object Recognition System Design in Computer Vision: a Universal Approach

**Andrew Gleibman**

Sampletalk Research, POB 7141, Yokneam-Illit 20692, Israel
www.sampletalk.com

**Abstract.** The first contribution of this paper is architecture of a multipurpose system, which delegates a range of object detection tasks to a classifier, applied in special grid positions of the tested image. The second contribution is Gray Level-Radius Co-occurrence Matrix, which describes local image texture and topology and, unlike common second order statistics methods, is robust to image resolution. The third contribution is a parametrically controlled automatic synthesis of unlimited number of numerical features for classification. The fourth contribution is a method of optimizing parameters *C* and *gamma* in LibSVM-based classifier, which is 20-100 times faster than the commonly applied method. The work is essentially experimental, with demonstration of various methods for definition of objects of interest in images and video sequences.

## 1 Introduction

Our primary concern is application of pattern recognition methods in computer vision, so we start with a brief description of related concepts and state of the art in the field.

Pattern recognition methods and techniques are described in many sources. The following description from Wikipedia, slightly modified here, clarifies the concept of *classifier*, see also [1]: *Supervised learning is the machine learning task of inferring a function from supervised training data. The training data consists of a set of training examples. Each training example is a pair, consisting of an input object (typically presented as a vector in some vector space) and a desired output value. A supervised learning algorithm analyzes the training data and produces an inferred function, which is called a classifier. The inferred function should predict the correct output value for any valid input object. This requires the learning algorithm to generalize from the training data to unseen situations in a "reasonable" way.*

One of the most successful supervised learning techniques in the latest decade is SVM [1]. Probability assessment for the predicted values (classes) is typically calculated using the Euclidean distance of the classified vector from the hyper-plane, which linearly separates training objects of different classes in a specially constructed vector space. Each input vector consists of numerical values, corresponding to certain numerical features of the object. When dealing with images or image fragments, the numerical features typically contain information about color, contrast, standard deviation of some area around the analyzed image point, second order statistics [3-5] for evaluating texture, general shape-related characteristics, presence of specific shapes, characteristics of image morphology, topology etc. While we are generally speaking of classification assuming the discrete class tags of the classified objects, we may also assume numerical values, such as real numbers, as the class tags.

The following terms are applied throughout this text in the following meaning, unless the context suggests another connotation.

*Sample:* A small, typically circular, area around an image point. This area is being analyzed and marked either by the user in *training phase* (see below), or automatically by the *classifier* in *image mapping phase*. For a given training phase and classifier, the shape and size of samples are considered constant. In our experiments, the radius of samples is typically 30-50 pixels where the image size is around 1 megapixel. In some cases, two or more concentric circles or rings can be applied simultaneously.

While sampling a part of the *object* for subsequent analysis in *training phase*, the user-tutor of the classification system is warned that the system can make a decision about presence of the object based only on this sample, without the image context outside of the sample. So, a sample should contain some distinctly visible elements, characteristic for the object or situation in question, which can be used for subsequent identification of this object or situation among other objects or situations.

Examples of circular samples and sampling sessions are given in our figures below. For specific applications, the shape of samples can be made rectangular, elliptical etc.

***Object:*** The concept of *object* here is intentionally made as independent of applications as possible. The object is considered a plurality of *samples* that have some common meaning for the user. The samples should contain fragments of some visible objects of interest (in a common sense), or display special situations of interest presented in the scanned scenes. The primary goal of *image mapping phase* is classification of *samples*, presented in the image, as belonging or not belonging to objects of specific classes.

This universal view of objects has many advantages, which are discussed in more detail below.

***Training phase:*** A session, in which the user defines class names (tags) for some *objects* and provides the system with tagged *samples* for subsequent automatic generation of the *classifier*, which will be applied for classification of similar samples in *image mapping phase*. The user is typically informed that so chosen samples should present distinct parts of the object, which are recognizable without any additional context. Training phase is typically applied iteratively, allowing the user to develop and correct *training set* in order to achieve better classification results.

The conclusive part of training phase is initiation of a special machine learning mechanism for generating the *classifier*.

***Training set:*** Set of *samples* with class tags.

***Feature vector:*** Vector of numerical or Boolean values, produced from a *sample* by special numerical feature algorithms analyzing the sample. Design and implementation of numerical feature algorithms for the multipurpose system are discussed in more detail below.

***Image grid:*** Image positions (points), placed with some constant step in horizontal and vertical directions. In *image mapping phase* such positions are typically considered centers of *samples* for automatic classification (mapping) by *classifier*. Examples of image grid positions are given in our figures below.

***Image mapping phase:*** Supplying *samples*, defined by *image grid* points, with class tags automatically by a *classifier*. So formed class tags are typically accompanied with probability assessment values calculated by the classifier.

***Classifier:*** A procedure, which automatically finds the most probable class tag using a *feature vector* calculated from a *sample* by special numerical feature algorithms. A classifier is typically created by a special machine learning technique using *training set* and the numerical feature algorithms.

The class tag is typically supplied with the probability assessment value. Furthermore, for every *sample,* the classifier provides the probabilities of this sample belonging to all other defined classes.

## 2 The Objects in More Detail

The concept of object, which is being defined and analyzed through *samples*, is central in this work.

In most image understanding systems, image fragments that contain the object being analyzed are extracted by some pre-processing procedure. For instance, in motion detection systems the objects are typically extracted using some image registration technique: two consecutive frames of a video sequence are registered (placed one onto another so that the frame shift is compensated), and the related image difference is then considered the object for subsequent analysis. Another example: in printed text recognition systems, the objects (letters) are typically extracted using some image projection techniques followed by morphological analysis, blob analysis, water-shed transform and other transformations. Then image fragments, which contain separate strokes or letters, are fed into the classification system for recognition of alphabetical letters.

Our concept of object is essentially different. This concept comprises "all image fragments, which are similar to those sampled by the tutor", while the similarity is checked for *samples* containing the tested points. Consider the positive and negative aspects of this approach.

*Positive aspects*:

1. The user can define, and then discriminate and automatically locate very complex objects and situations such as a pedestrian on a road, a vehicle from a large class of vehicles, a building construction, vegetation, an instrument, a pre-contact or contact situation, a human hand or iris, an assembly etc. (Object examples of this kind are given below). The nature of such objects and situations is unrestricted by definition. There is no need for a sophisticated image pre-processing procedure.

2. With modern computer power, training and subsequent recognition is possible for highly complex objects and situations, which may be defined by tens of thousands or even millions of image samples.

3. The object being analyzed in *image mapping phase* can be presented only partly. For example, a vehicle can be detected just by locating the vehicle tire or identification plate, a human can be detected just by detecting human face or hand. So, the system is robust in various screening situations, where only parts of the object of interest are visible.

*Negative aspects:*

1. For a successful training, the system may require a very large quantity and variety of sampled points. Our approach to efficiency of automatic training helps to cope with this requirement (see *Avoiding Exhaustive Search in Training of the Classifier* in Section 5).

2. Samples can be ambiguous or misleading, and the tutor needs to be sure that *he* would recognize, both in training and conceivable mapping phases, any such sample without the context outside of the sample boundary;

3. The system can be vulnerable to image resolution, scaling, illumination, and to the variability of neighboring contexts, not belonging to the desired object but belonging to *samples*.

Compensation of these negative aspects is discussed below. The tutor has the possibility to efficiently correct and adjust the training set according to his recognition needs and to the presence or absence of unique characteristics of the desired object in samples. So, a training session typically consists of repeated iterations. The mentioned efficiency of automatic training is a great asset here.

**Image mapping at grid points.** Image mapping should be considered a framework for *detecting objects of interest through classification* (cf. Fig.1). An image may contain many objects of interest, so we apply the classifier in all image points placed on the image grid. The user defines the grid resolution and the radius of *samples*. Classification results on the grid points can be presented using a color-highlighting of object classes of interest. Examples of color-highlighting are given in figures below.

**Multi-object location and detection.** Consider Fig. 2-11. Image mapping phase provides a simple dialog, in which the user can observe the classification results. Using a drop-box dialog with defined class names, the user can choose a class and observe only results of detecting objects of this class. In Fig. 6 such classes are "Contact Situation" and "Human Fingers" respectively. Another option is simultaneous observation of several classes marked by different color marks on a single image. In this way *image mapping phase* can be immediately applied not only for classification but also for detecting and locating objects of interest of different kinds, as well as for raising alert messages in special cases where objects of interest are found or not found.

Fig. 6 (*right*) and 7 (*right*) demonstrate the situation where intrusion of a human hand is detected and reported. Fig. 8 (*right*) demonstrates the situation where the absence of open eye is reported. So, the computer vision task of detecting the presence, absence, or the dynamics of special objects or situations, is delegated to the *classifier* via the image grid.

# 3 Discussion: Minimization of formal knowledge applied in cognitive and machine learning systems

The presented *universal approach* should not be considered an attempt to build everything related to pattern recognition in computer vision using a single scheme. Like a partially ordered set can have many maximal elements in mathematics, there can be many multipurpose system architectures, any of which is suitable for creation of a wide range of custom applications of certain kind. Here we relate *universality* to a range of applications, where objects of interest can in principle be learned and detected through *samples*.

A similar concept of universality was developed in our research in First Order String Calculus (FOSC) [6, 7]. FOSC is a generalization of classical first order logic. In this generalization, unrestricted strings are applied instead of logical terms and predicates. Variables are created by string alignment: given aligned string examples, their non-matching parts can be replaced by variables. In this way strings are *generalized* and may contain variables and some nesting structure. Variables in FOSC are treated similarly to object variables in First Order Predicate Calculus (FOPS). This kind of logic provides a way to build natural language processing applications immediately *from examples* without the concept of formal grammar [7]. Formal theories and algorithms are extracted from unrestricted text data in fashion of *inductive logic programming* [9], but without the need to invent logical predicates. In a way, this is the simplest formal framework for modeling cognition and human intuition. The usage of *a priori formal knowledge* is minimized up to an asymptotic limit, which for the case of 1st order logic is Robinson's resolution principle [11]. All formalisms, necessary for specific applications, are generated automatically by analysis and generalization of sample data.

With this minimization, Church-Turing thesis [10] is reformulated as follows: *Every possible computation can be carried out by alignment and matching of suitable data examples*, with a specific definition of *alignment* and *matching* as the universal algorithm definition means applicable in a machine learning framework ([6], p.110).

The resolution principle is applied in FOSC as the main inference engine in the same way a *classifier* is applied as the main inference engine for reasoning about images in our present research. The present research expands the *a priori formal knowledge minimization* framework for working with image data. The minimization leads to a simple and nearly-universal set of numerical feature algorithms for classification of unrestricted images. Specifically, our experiments demonstrate the asymptotic simplification of the numerical feature algorithms, which can be combined through SVM [1, 2] and known feature selection methods [12, 13] into a wide spectrum of image classifiers based on unrestricted training data.

A similar minimization for extraction of astronomical equations immediately from observational data is described in [8].

**Universality of the pattern recognition component**. So, the pattern recognition component of the system is intentionally designed as universal as possible for a wide class of applications related to computer vision.

Note, however, that universal numerical features, suitable for all kind of unrestricted patterns, do not exist. Some methods are good for discriminating textures and color variations of the objects, while other methods are better for recognition of object forms and edge characteristics. Nevertheless, there exist powerful methods, suitable for a wide spectrum of application areas. Among these methods we should mention those based on second order statistics [3-5] and Fourier, Gabor and Wavelet transforms, with calculation of power spectra in specific frequency bands.

This approach assumes using the most general numerical features, suitable for possible applications in a wide range of areas, where the feature combinations ar suitable for definition of classes in more specific areas. The number of numerical features, actually applied for a specific pattern recognition task, is controlled by known feature selection methods [12, 13].

**Architecture and customization of the universal system.** The architecture and the numerical feature algorithms of the presented system are influenced by our research presented in [6-8]. For building a custom object detection system, based on this approach, the following scheme is applied.

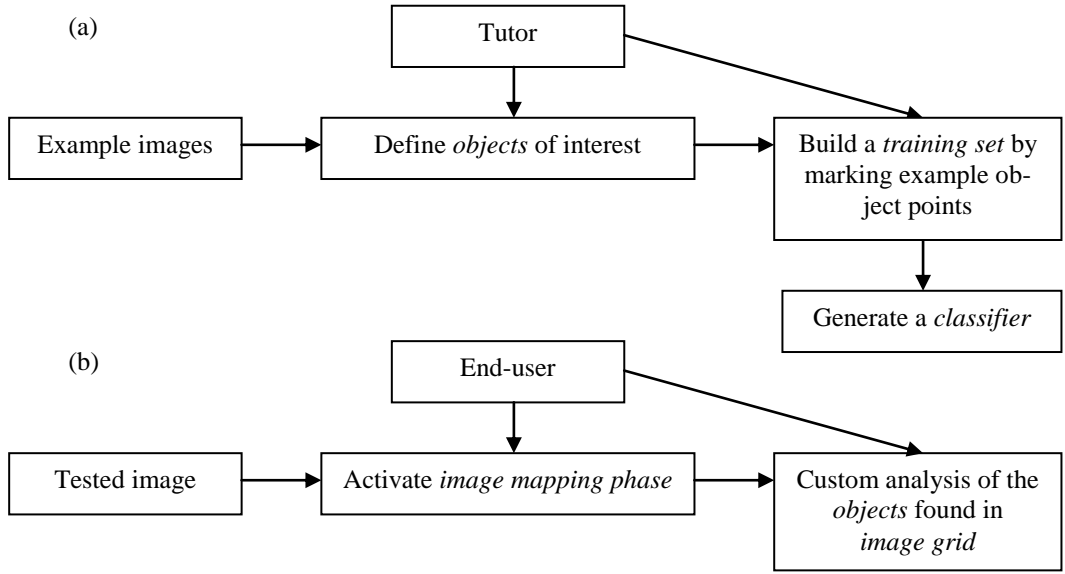

Fig.1. Delegating object detection task to a *classifier*: a) *training phase*, b) results of *image mapping phase* are applied for custom reasoning about presence, absence, location and combination of objects of interest.

Using this scheme, the building of a custom object detection system is reduced to a training phase (Fig.1, a) and doing a custom analysis of the classification results in grid points of the tested image (Fig.1, b). For instance, location of the object of interest (Fig 4, *top right*) or presence of the object of interest (Fig. 6, *Fig7, right*) can be reported. Other custom situations may include counting samples of the objects of interested (Fig 8, *left*), detecting the absence of the object of interest (Fig. 8, *right*). In some situations, a more sophisticated analysis of the image mapping results can be applied. For instance (Fig. 10, right), the detected points of class Pupil are clustered into two clusters, and then a calculation of the shift of the cluster centers relating to the human face is applied for calculation of the gaze direction. In our examples, the objects of interest (pupils in Fig. 10 and 11) are successfully located in spite of the complex natural-light illumination with light reflections.

**Comparison with Viola-Jones object detection framework.** Viola-Jones's approach [14] is essentially based on evaluation of rectangular image regions for a set of special rectangles, which feature the object of interest. In the meaning of *universality* discussed above*,* this approach is universal too. Using an advanced machine learning scheme (AdaBoost), it also applies a possible minimum of a priori formal knowledge.

In our framework, however, the objects of interest can be more complex. An *object* can represent a complex and synthetic concept, which includes essentially different sub-classes. As an example, consider the object "Building Construction" in Fig. 4. *Samples* of this object may include various roof elements, wall fragments, chimneys, windows, fences, wall angles etc. Definition of so complex objects in Viola-Jones's framework seems to be problematic.

In our framework, the results of *image mapping phase* can be analyzed for various purposes, as shown in Fig 1 and examples 2-11. Viola-Jones's framework is more similar to a black box, which does not allow the user to analyze the plausibility assessment of object detection in individual image points.

While we are concentrating on object classification and detection at exact image grid positions, the Viola-Jones's framework work with entire image regions. Additionally, training time in Viola-Jones's algorithm is typically much larger than training time in our case (cf. Section 5).

# 4 Universal Numerical Features: Texture, Topology, Gray Level, Power Spectrum

Color images, presented in our experiments (Fig. 2-11), are considered triples of R, G, B components, each of which is analyzed as a gray level image. The numerical features for classification of every *sample* are extracted from all three bands and then joined into a single *feature vector*.

Consider the analysis of gray level images in the framework of universality.

**Gray level factorization.** Usage of second order statistics [3] depends on the range of gray level values applied for building Gray Level Co-occurrence Matrix (GLCM). For example, the common set of 256 values (0-255) results in a 256*256 matrix. For texture classification, special statistical values (homogeneity, entropy, contrast etc.) are typically extracted from this matrix [4, 5].

The result of using such values in classification algorithms essentially depends on the applied set of gray level values. Although this observation seems to be trivial, it can be neglected, and GLCM (as well as other matrices introduced below) will be applied not in the most effective way.

Our approach suggests a numerical parameter for factorizing the standard set of 256 values 0-255 into a smaller set, typically 0-26 or smaller.

This factorization is justified as follows. An ordinary human eye cannot discriminate gray level values whose difference is less than 10 gray levels in the standard 0-255 scale. Additionally, the distortion of brightness, produced by optical scanning devices, often surpasses this difference. So, the GLCM-based mechanism for modeling human eye recognition needs not more than 0-26 gray levels. Additional benefit of using a smaller grey level scale is computational efficiency and memory usage, which is essential when a special hardware is applied. Considering the efficiency of hardware implementation, gray scale factorization 256/8 or 256/16 is preferable, so the optimal number of resulting gray levels will be 32 or 16.

**Orthogonal gradient co-occurrence matrix.** Gray level co-occurrence matrix (GLCM) [3,4] takes into account neighbor pixels in one or several directions. As an addition to GLCM, we introduce a special *Orthogonal Gradient Co-occurrence Matrix* (OGCM). For every pixel of the analyzed area, two values *g1, g2* of image gradients in orthogonal directions are calculated. This can be done either for a pair in (SN, EW) directions, or both for pairs in (SN, EW) and (NW, NE) directions, as we do in our experiments, or for more pairs of orthogonal directions. The OGCM matrix is then built, similarly to GLCM, by incrementing element (|*g1*|, |*g2*|) of the matrix for every analyzed pixel.

This matrix provides additional features of form and texture. In a way, it characterizes a 3D form of the objects presented in a 2D form. In some of our experiments OGCM matrix was applied for generating second order statistics feature values similar to that typically extracted from GLCM. Application of Gray Level-Gradient Co-occurrence Matrix [5] is recommended too.

Important feature of GLCM and OGCM is low calculation complexity, which is linear in the number of pixels of the analyzed region. This is essential when a special hardware implementation is assumed.

Our experiments show that when SVM is applied with a RBF kernel, numerical feature values of homogeneity, entropy, contrast etc., typically extracted from GLCM, can be avoided without any loss of the system functionality. Instead, the matrix elements themselves can be applied as the numerical feature values for classification. This observation is consistent with our framework of *universality* and *minimization of a priory formal knowledge*, described in Section 3. Complex texture concepts of homogeneity, contrast, entropy, uniformity etc. turn out to be redundant. The machine learning system generates the necessary texture description automatically based on elements of GLCM and OGCM matrices as the numerical feature values.

**Shape recognition.** For recognition of specific shapes, the numerical features, related to texture analysis, are typically strong in high frequencies but weak in low frequencies.

Using SVM with RBF kernel, one does not need to explicitly define specific shape models such as bars, cross-bars, circles, triangles, arrows, arcs etc. The power of modern machine learning techniques allows one to provide the *universal* system with *samples* of such objects and to generate the recognizer of specific shapes automatically.

This, however, may require the ability of the system to catch low frequencies of the analyzed images. To address this issue, we do a 2D FFT of the *samples*. Then, the values of power spectra in low frequency ranges are immediately applied as the numerical feature values for classification. Another, more efficient option (also applied in our experiments) is using power spectra of 1D wavelet transform of several straight lines, which intersect the *sample* in different directions. These values are accumulated into a single row of numerical feature values, which are then applied for shape representation.

**Gray level-radius co-occurrence matrix for samples.** Classification, based on texture (GLCM, OGCM) and power spectrum-related feature values as described above, is typically vulnerable to changes in image resolution.

For a more robust presentation of image topology we introduce a special Gray Level-Radius Co-occurrence Matrix (GLRCM). In order to build this matrix, a *sample* is considered a set of concentric rings around the analyzed image position. The rings are constructed with sequential radii of external circles *w*, *2w, 3w, …, Nw=R*, where *w* is some constant width for all rings and *R* is radius of *samples*. For any of the rings, a factorized gray level histogram is calculated, where *G* is a maximal gray level value. Then, the matrix of *N*G* size is built by setting value of the *(n, g)* matrix element to the number of pixels of gray level *g* in ring number *n*.

In most our experiments, the classification feature values, taken immediately from this matrix, lead to better results than those based on GLCM and OGCM. Indeed, GLRCM contains information about local image texture in combination with local shape characteristics independently on the image resolution.

**Parametric numerical feature synthesis.** Feature extraction methods, described above, are essentially parameterized by the following parameters:

1) Degree of gray level factorization (which affects dimensions of GLCM, OGCM and GLRCM);
2) Ring width *w* (which affects dimensions of GLRCM);
3) Radius *R* of *samples*, which affects the number of accumulated values in the matrix elements;

Variation of these parameters substantially affects the number of numerical feature values, applied in classification algorithms, independently on feature selection methods. Additional parameters are feature selection thresholds for correlation, entropy, and other statistical methods related to *mRMR* [12, 13] etc).

In our experiments, special system architecture is applied, which allows variation of such parameters independently of other system components. This leads to *a controllable synthesis of unrestricted number of numerical features for classification*. In order to achieve better classification results for a practical work in *training phase* and *image mapping phase*, this number can be easily controlled.

The number of so generated numerical features for classification, applied in our experiments, varies from a few dozen to a few thousand.

## 5 Application Aspects of the Multipurpose System

**Circular areas for sampling image points.** The treatment of objects of interest through *samples* requires a special interface for object definition. When creating a training set for generation of the classification algorithm, the user can observe training images, choose and mark sample points, and then tag them as belonging to the defined classes. The system contains a simple dialog, which displays a circular area around the observed point. The dialog provides simple means for tagging samples with class names. Again, the tutor is instructed that the system does not take into account image areas *outside* of the circle, so he should choose those samples that contain enough information for classifying the center of the *sample* as belonging to the object in question. Examples of this interface in training phase are provided in figures below.

**Correcting the training set.** After some version of the classifier is built, the system provides the tutor with the possibility to observe the results of classification on the objects, which belong to the training set. Image fragments, which still are not included into the training set but may possibly be included, are analyzed too.

For every defined class, and with every desirable probability value provided by the classification system according to [2], the tutor can observe, and then manually correct the classification results obtained automatically. In this way the user expands or corrects the training set for the next iteration of training phase.

A simple editing dialog is defined for this correction. In this dialog, the user is presented with a mapped image. Using a drop-box with a list of defined class names, he can choose a correct name and then click the object whose class name should be corrected. As the result, a new named *sample* will be inserted into the *training set* for subsequent re-training.

**Removing ambiguous training samples**. Removing wrongly tagged samples is possible by a similar method. Image mapping phase is activated for an image. The user observes the detection results for different classes and analyzes wrongly classified objects. If a classification error persists, the user can refer to the training images where the respected classes are sampled and tagged, and correct the class tags for subsequent re-training.

**Image scale treatment.** The analyzed scenes can be scanned from various distances so that the objects, contained in the scenes, are presented in various resolutions and may take image parts of various sizes.

Without a special treatment, the classifier, trained in one scale and resolution will not work well in another scale and resolution. The situation, where the analyzed objects can be scanned from different distances, should be reflected in *training phase*. This may require a large number of manually analyzed images of training scenes and situations.

Another view of resolution is related to the object shapes. For instance, in a detailed resolution, an image of a tree may display texture of leaves and branches but miss the general shape of the tree. In a less detailed resolution, the GLCM and GLRCM techniques may catch the general contour of the tree. In this way, both leaves and contour of the tree are featured. Two or more sizes of *samples* can be defined simultaneously, and the numerical feature values are produced using both sample sizes. This technique is typical for a multi-resolution analysis.

In this way, the finer resolution is analyzed for a texture, while the less detailed resolution may present the object shape characteristics.

**Skipping non-informative image fragments.** In *image mapping phase*, some image fragments may contain no useful information for the analysis so that the expensive process of classification is meaningless. A special information mask can be applied for skipping non-informative areas. In our experiments, this mask is created using Niblack binarization [15] with subsequent edge detection. Other methods of avoiding non-informative image areas can be applied too.

**Avoiding exhaustive search in training of the classifier**. Pattern recognition component of the system is currently based on LibSVM library [2]. This library builds a classification algorithm given a training set of classification examples, presented in some form of object-feature matrix with numerical class tags for every object.

In building the classifier, the usage of this library essentially depends on the choice of special numerical parameters. Specifically, the choice of non-linear transformation (kernel function), as well as of the penalty parameter $C$ of the applied error term should be made. Kernel function may include two parameters ($d$ and $gamma$ if a polynomial kernel is applied, $r$ and $gamma$ if a sigmoid kernel is applied), one parameter ($gamma$) if a radial basis kernel is applied etc. Together with parameter $C$, the values of two or more numerical parameters should be chosen and optimized.

Today a viable theory for choosing optimal values for these parameters does not exist. In paper [2], a simple "greedy" algorithm is proposed for funding a near-optimal solution. For example, if a radial-basis kernel is applied, 20 possible values for parameter $gamma$ and 20 possible values for parameter $C$ are analyzed in a grid of $20*20=400$ combinations. A trial classifier is built for every such combination, and the best pair of values $gamma$ and $C$ is then applied in final training of the system. A cross validation technique is applied for the assessment of every so generated classifier.

The negative side of this $20*20$ grid exhaustion algorithm is efficiency. For instance, when a training set contains 10.000 or more of objects with 5-10 defined classes, generation of the final classifier by LibSVM may take 10 or more hours on a standard PC computer with 2.6 GHz processor. In case of three parameters (3D grid exhaustion) such search is impractical or impossible at all.

Our solution to this problem is *a random access to the grid points of parameter values* instead of the exhaustive search (not to be confused with *image grid points*). Furthermore, the grid can be avoided entirely. We can simply generate random values for the parameters in appropriate ranges. In our experiments the system comes to a nearly-optimal solution, similar to that described in [2], 20-100 times faster using the same computational resources. This solution can be applied for a 3D or 4D parameter search too. During the search, the classifier assessment value (total error rate) is displayed in a system log, so the user can stop the search if a sufficient precision/accuracy is already found. Otherwise, the user can stop the search if he observes that the system cannot generate any viable classifier after 10-20 trials, instead of 400 or more proposed in [2]. The latest case typically indicates that the training set is inconsistent and should be corrected or entirely redesigned.

In the following table, the classifier training time is compared for the grid exhaustion method [2] and for our method of random access to the grid points of parameter values. The pair of parameters *gamma* and *C* is optimized here. The qualities of the classifiers, built by both methods, are similar (97-99% accuracy in most cases). In this experiment, a single 2.6 GHz Intel processor is applied.

| *No. of defined object classes* | 3 | 3 | 3 | 12 |
|---|---|---|---|---|
| *No. of applied classification features after feature selection* | 24 | 314 | 26 | 942 |
| *No. of object samples in training set* | 2200 | 4200 | 290 | 8620 |
| *Classifier generation time using the exhaustive grid search [2], minutes* | 74 | 320 | 42 | 740 |
| *Classifier generation time by the random access method, minutes* | 3 | 12 | 0.5 | 21 |

Table 1. Comparison of grid exhaustion and random access methods for optimizing *gamma* and *C*

This approach not only makes the training more efficient. In many cases it allows correction and development of very complex training sets that otherwise cannot be developed at all. Indeed, if the tutor of the system has to wait many hours for evaluating every small correction of the training set, his teaching process is drastically limited. On the other hand, if any trial training takes only several seconds or minutes, he can perceive and better control the building of the classifier. He can easily correct wrongly defined classes and wrongly tagged training samples, observe and remove ambiguously tagged objects from the training set, add/remove classes, tag additional samples etc.

**External interface for the pattern recognition component of the system**. Additional components of the system contain basic functions for operation with the results of *image mapping phase* on the scanned images or scenes. Among such functions we should mention:

– Rising alert signals about special objects and/or situations and/or class combinations presented on analyzed images;
– Behavior checking concerning the dynamics of certain classes in certain locations;
– Behavior control guided by the behavior checking;
– Detecting presence or absence of special objects and/or situations either in the whole image or in specific locations;
– Locating special objects and/or situations;
– Counting the number of special objects and/or situations either in the whole image or in special locations;
– Counting the number of grid points belonging to objects of special classes either in the whole image or in special locations.

Most such functions analyze the probability estimations, provided by *classifier* in every *image grid* point. Using such estimations, the user typically can define a trade-off between the classification error rate and the rate of covering the observed objects.

**Working with frames of a video sequence.** Video sequences typically provide some additional information, useful for image classification, especially in situations where the objects of interest are moving or inserted otherwise. In such situations, the static background may aggravate the image mapping phase of the system. Indeed, since *samples* may contain fragments of the moving object of interest, as well as different fragments of background, the background affects the calculation of numerical features for classification.

This problem can be resolved by either of the following two methods:

1) Learning the moving objects in all possible background contexts;
2) Removing the static background from the video frames.

Method (2) is preferable where the background is complex. Various methods can be applied for removing a static background. Our preferred method of intrusion detection is based on registration of a current video frame with a reference frame, followed by removing all registered point pairs that belong to well-correlated image fragments of certain size and form.

**From smart camera to intelligent vision system.** The market of smart cameras has flourished in the latest decade [17]. Typically, a smart camera facilitates tracking of moving objects and has functions for recognition and evaluation of special objects and situations: vehicles, pedestrians, road situations, defects in a production line, vehicle number plates etc. Every such system is typically based on specially built intelligent image processing and pattern recognition algorithms. Examples of such systems include an optical mouse, which detects motion direction with high speed and precision, SmartEye [19], which measures either the point of gaze or the motion of human eye relative to the head, MobilEye [21], which detects various objects and situations on the road and warns the driver.

Pattern recognition algorithms of such systems typically are specialized and embedded in a special microchip. The cost of creating new microchips, however, leads to a decline of specialized chip manufacturing systems [17, 18]. This is compensated by companies, which produce only intellectual components of the chips – System on Chip (SoC) IP cores. For a small or medium-size company, designing one or more IP Cores and ordering the manufacture of related microchips in already existing FAB is more practical than doing everything in a single site.

The aspect of our approach, which is useful here, is development and standardization of the pattern recognition component, which is the heart of all such systems. We develop a trainable multipurpose system for pattern recognition, which provides a method for manufacturing specific devices working with images and video sequences. Besides using the most universal and powerful pattern recognition methods, we create a special interface for connecting such algorithms with other system components, as well as a special interface for connecting all those components with the external world. Producers of specialized systems, such as security surveillance systems, vehicle safety control systems and industrial line control systems, will be able to adapt this single universal core instead of designing special algorithms for their purposes.

The described system not only can be trained and viewed as a smart camera. It is really an intelligent vision system, which may contain one or more cameras, powerful image processing and pattern recognition algorithms, and specially designed interfaces for teaching the system how to apply these components in a specific environment.

**Video sequence understanding: real-time (online) vs. offline processing.** Our view of an intelligent vision system comprises image understanding ideas in combination with hardware organization and interface ideas. The system is capable of performing such tasks in less demanding environment, concerning efficiency, where a conventional PC is connected to a video camera, or where otherwise collected video sequence is fed into a computer for an off-line analysis.

Large volume video data is widespread today. Typical scenario for collecting data is logging (DVR recording) of road or industrial process situations. The logging may include additional video or non-video data, GPS-receiver info, speed, temperature, gas consumption, audio info etc. [22-24]. An intelligent vision system should support detection, location, evaluation, tracing and counting special objects of interest in a large data log of this kind.

This task can be performed off-line by teaching the universal system how to analyze the log.

For teaching the system we support the following scenario. The user analyzes sample parts of a large video sequence and manually points some objects of interest: a certain person in a crowd, a special vehicle in a traffic, specific objects or defect types in a production line, human hands at a manipulator, electrical contact situations etc. The system learns such objects from the user's examples. Then it locates all such objects in the video sequence and informs the user for further analysis of this data.

A similar scenario can be applied for the analysis of video sequences in online situation. In this way, the system can be applied surveillance of an industrial process for a real time reporting about special objects and situations.

Additional tasks possible by training of the system: driving style evaluation (detecting and recording only those situations in which the driver violates driving rules) − for insurance companies; driving style correction (detecting driving errors and issuing recommendations) − for the driver; detecting singular situations in security systems or in production lines, such as intrusion or absence of hands or open eyes, danger of extra contacts or absence of necessary contacts when using electrical appliances, detecting alien objects, detecting a superabundance or a lack of special objects in the scene, detecting special instruments and/or objects in medical operational environment, marking, recording and reporting singular or dangerous road situations in real time, adding object recognition and scene evaluation functions to current DVR systems etc.The functionality for performing such tasks is based on the universal pattern recognition component of the system and on the described special alerting/controlling functions.

## 6 Examples

In Fig. 2-11 our experiments in detection, location and classification of typical objects and situations on images and video sequences are demonstrated. The experimental system includes a simple alerting/controlling language, which supports creation of specific interfaces for working with the pattern recognition component. Video frame examples are given along with the alerting messages, which expose presence or absence of specific objects and situations. Once more, the recognized objects and situations are not restricted by any predefined application area. Furthermore, the reaction of the system is not limited by just finding an object of a given class. The range of system reactions is wider: detecting motion, locating specific objects on still images and/or video frames of the scanned scenes, raising the alerting messages when specific objects *are* or *are not* found, calculating number of points belonging to detected objects of certain types (cf. Fig. 8 and 9) etc.

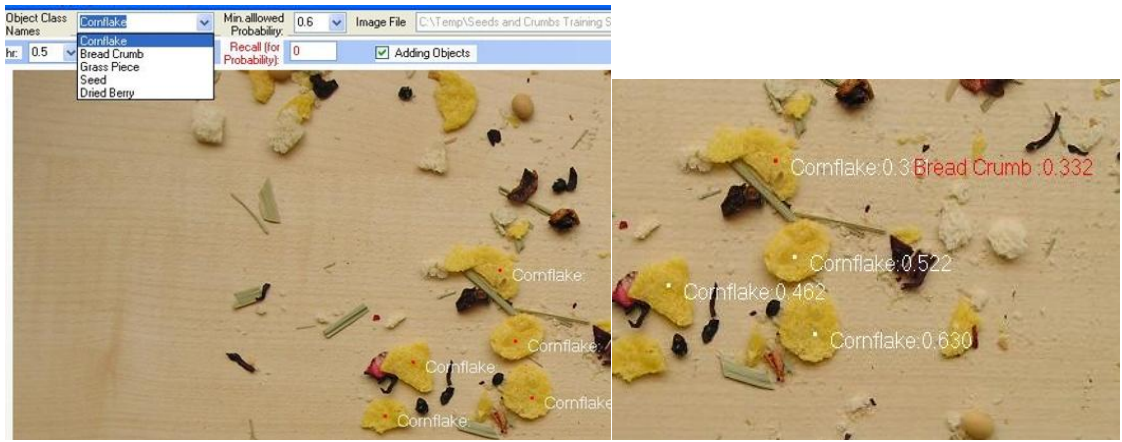

Fig.2. *Left: Training phase*. Choosing and marking objects of interest by class names. In this example, objects of class Cornflake are marked. *Right: Image mapping phase* applied for the training set objects. The numbers indicate the plausibility assessment values generated by the classifier. Red color alerts the user that the generated classifier made a wrong classification of a training object.

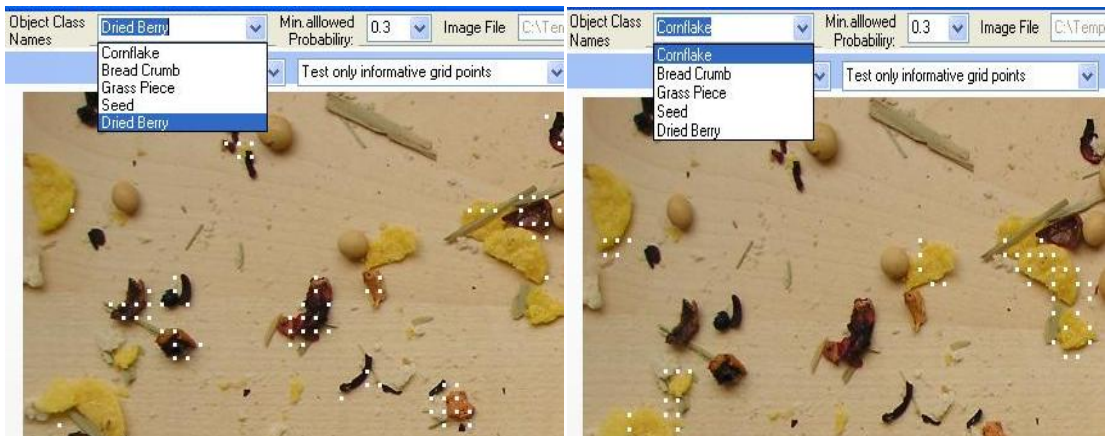

Fig.3. *Image mapping phase* applied for a new image not belonging to *training set*. White points indicate the grid positions where objects of classes "Dried Berry" (*left*) and Cornflake (*right*) are detected. The images display some misclassified and non-recognized objects of defined classes. Note that the system displays only those detected samples that have the plausibility assessment values equal or greater than the plausibility limiter defined in the dialog system (value 0.3 in our example).

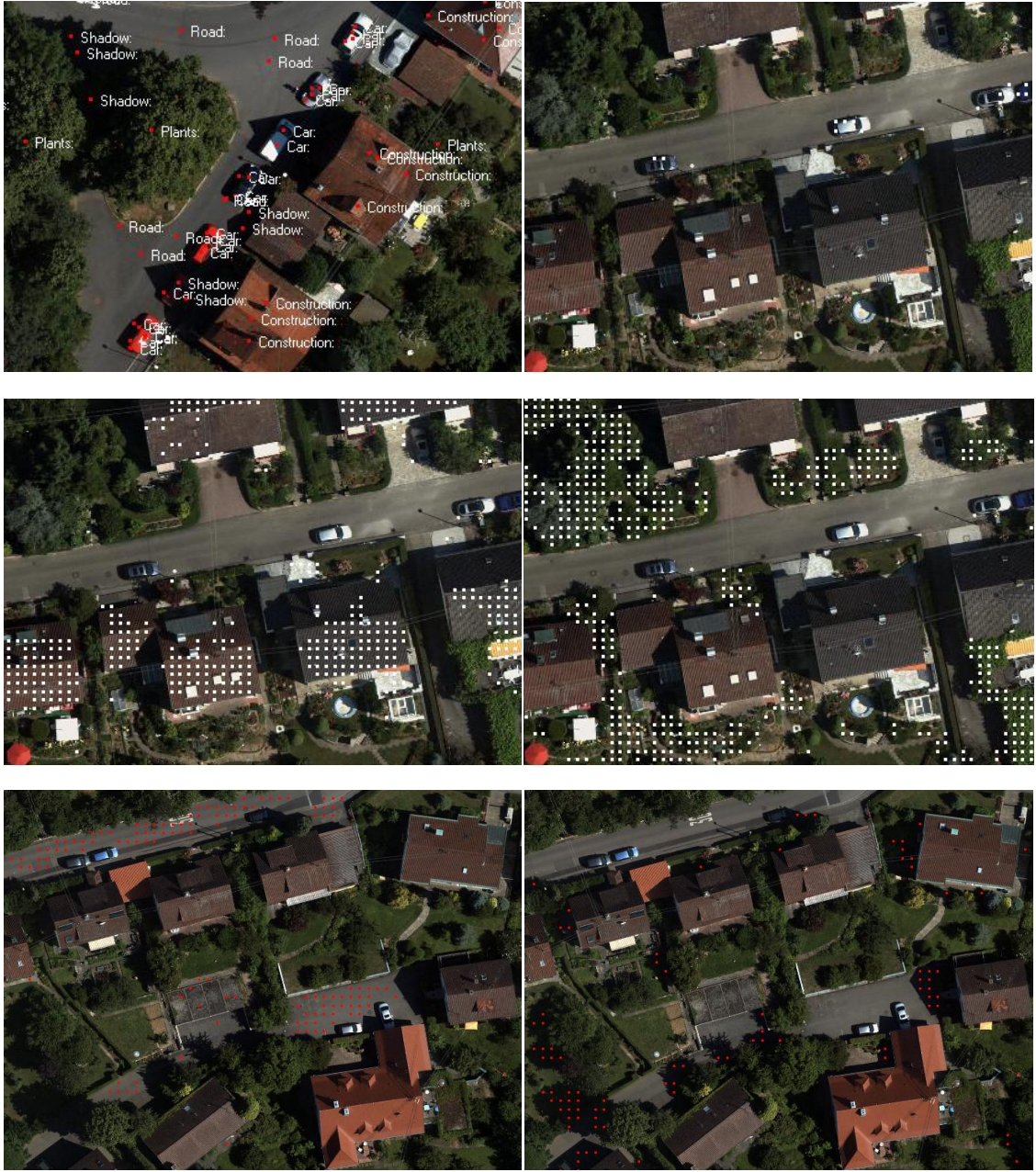

Fig.4. *Training phase* (*top left*) and *image mapping phase* for airborne images. The detected and marked *objects* of interest are vehicle (class Car, *top right*), building construction (*middle row left*), vegetation (class Plants, *middle row right*), terrain feature (*bottom, left*), shadow (*bottom, right*). White and red points indicate the grid positions where *samples* of these objects are detected. *Image data: Courtesy of VisionMap Ltd.*

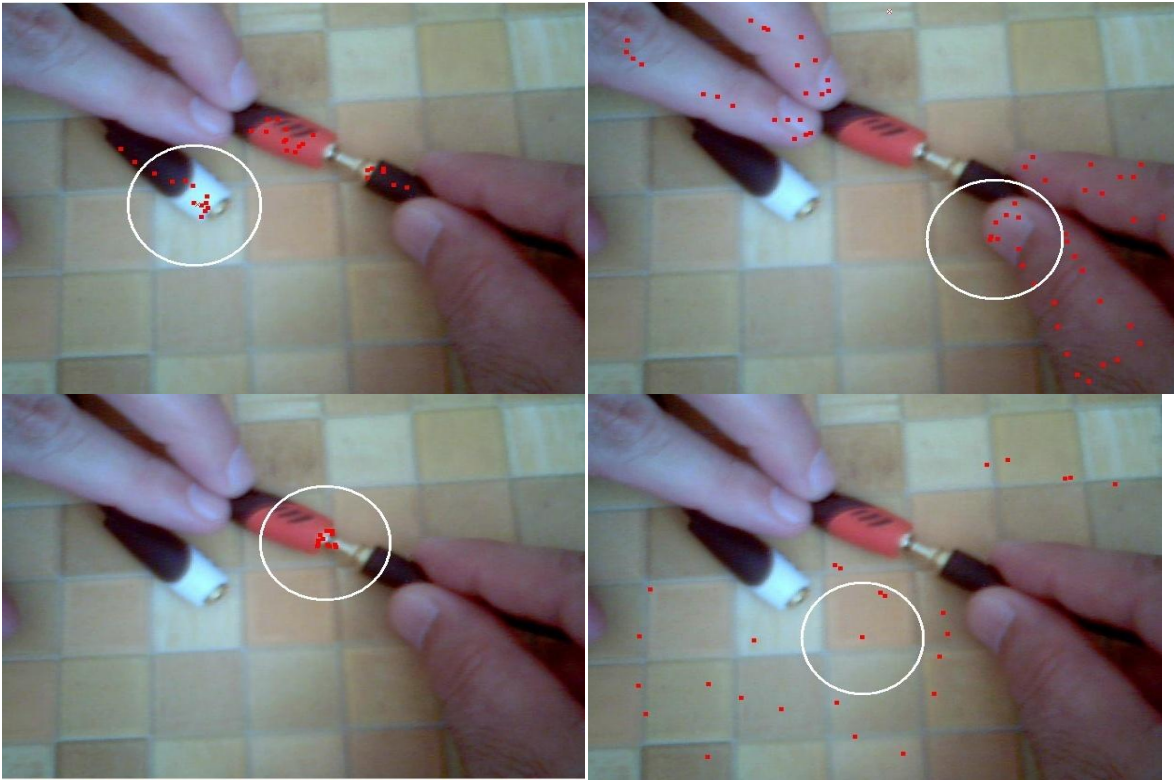

Fig.5. Sampling objects and situations of interest for subsequent classification in a video sequence. In t*raining phase*, the user is sampling parts of electrical appliances (*top left*), parts of human fingers (*top right*), contact situations (*bottom left*) and a background (*bottom right*). In i*mage mapping phase* similar objects and situations are detected automatically by the *classifier* (see the next image).

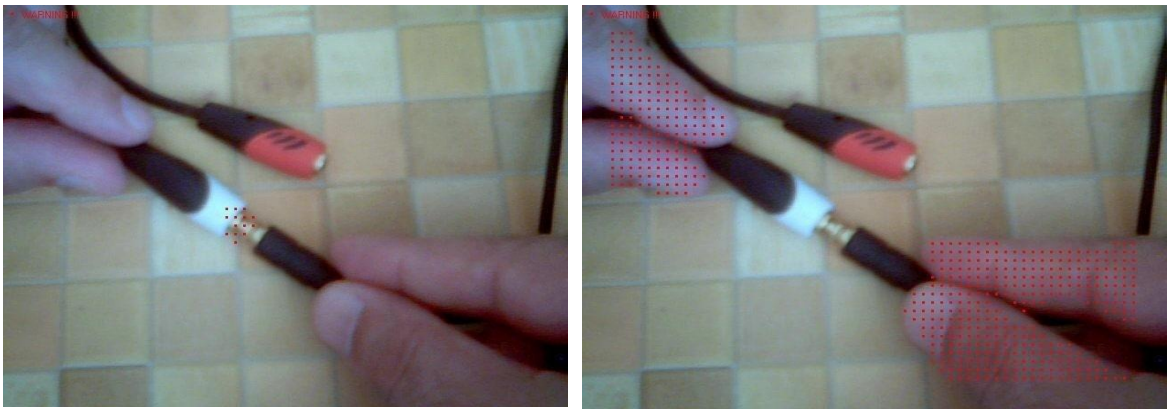

Fig.6. Detecting and raising an alerting message about a contact situation (*left*) and about intrusion of human fingers (*right*). The alerting message is displayed in the top left part of the video frames.
See  https://www.youtube.com/watch?v=Aj2cizRfaus  and  https://www.youtube.com/watch?v=LqMtNGjTqDI  for the live video demo of both scenes.

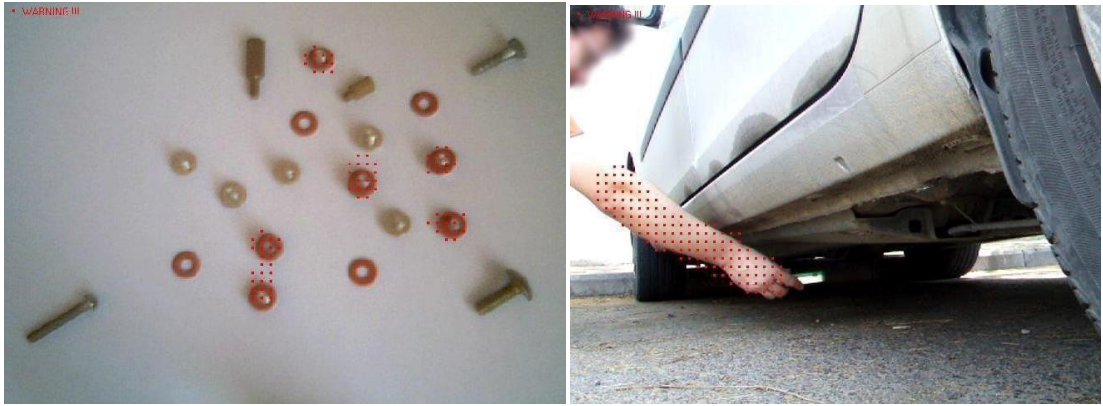

Fig.7.*Left:* Detecting assemblies among simple parts. See https://www.youtube.com/watch?v=dbZdUS6yHzw for a live video demo. *Right:* Detecting and raising an alerting message about intrusion of human hand in a security environment. The alerting message is displayed in the top left part of the image.
See https://www.youtube.com/watch?v=YfskAeAJCwI for a live video demo.

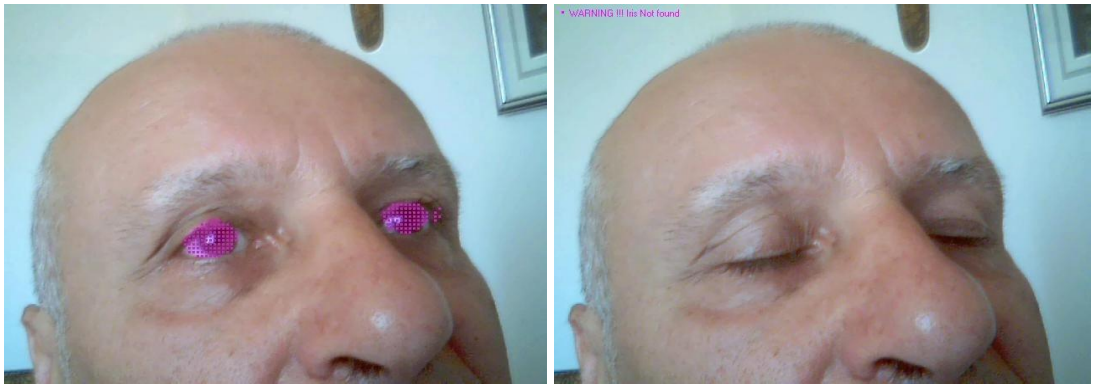

Fig.8.*Left:* Detecting human eye iris and evaluating degree of openness of human eye by counting the number of detected iris points belonging to the image grid. *Right:* Raising an alerting message when human eyes are closed, for behavior checking and control. The alerting message is displayed in the top left part of the image. See https://www.youtube.com/watch?v=yZBQ6cXOm4Y for a live video demo.

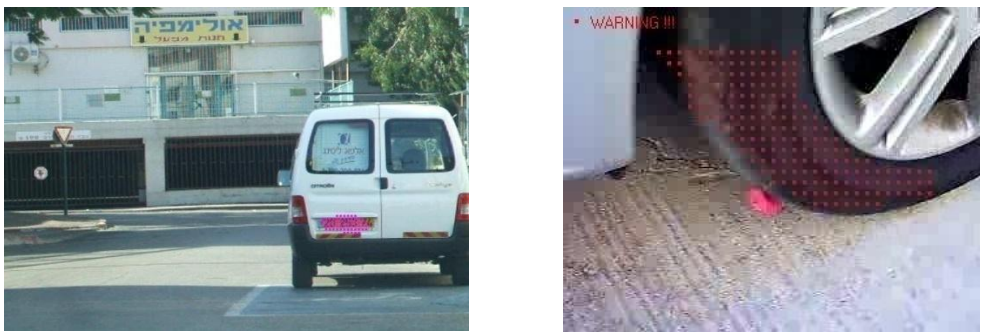

Fig.9.*Left:* Locating vehicle number plates for subsequent vehicle identification by OCR.
See https://www.youtube.com/watch?v=vTreTbSDaAQ for a live video demo.
*Right:* Detecting vehicle tires among other objects; evaluating the air pressure in tires by counting the number of detected image grid points of class "Tire" in the road-contact area.

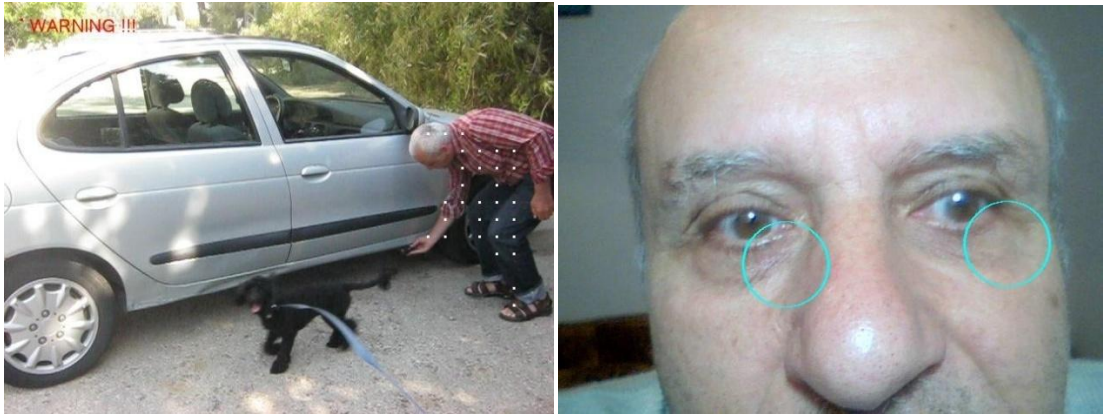

Fig.10.*Left:* Detecting intrusion of a human while skipping the animal in a homeland security environment. See https://www.youtube.com/watch?v=fbxuFqP2Q58 for a live video demo. *Right:* Gaze detection in complex natural illumination environment. See https://www.youtube.com/watch?v=9XH16k2gIUc for a live video demo.

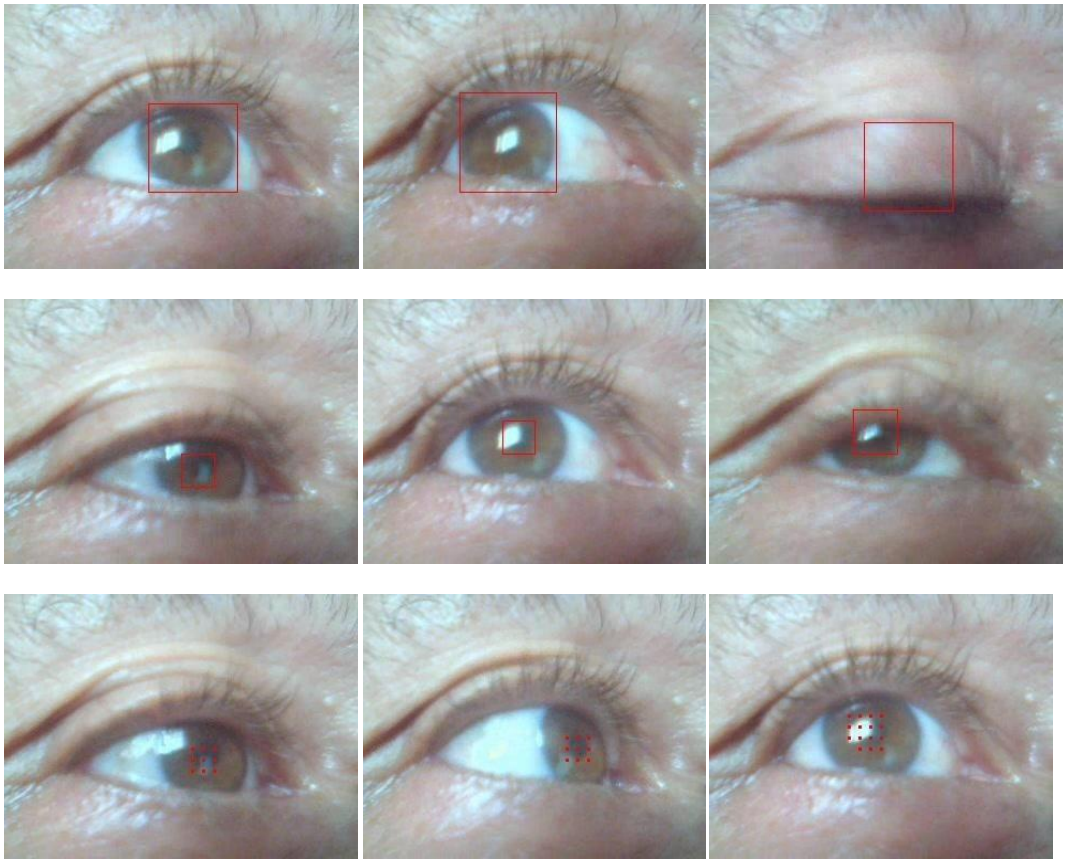

Fig.11. Detection of human iris (*top row*) and pupil (*middle row*) in complex natural illumination. When the eyelid is completely or partly closed (*right images of two upper rows*), the iris (pupil) position is found by an interpolation using consecutive frames of the video sequence. The *bottom row* shows actually detected points of class Pupil, which are applied for constructing the above enclosing rectangles and for the interpolation.
See https://www.youtube.com/watch?v=QMlSgYYcOac for a live video demo.

## 7 Conclusion

A specific concept of multipurpose system for pattern recognition in computer vision is introduced. Complex backgrounds, occlusions and image distortions are considered. The system contains pattern recognition algorithms, based on classical machine learning principles, which can be organized for performing so various tasks as raising an alerting message, locating and counting special objects, presented in the scanned scenes, detecting the presence, absence, or specific combination of objects, evaluating the dynamics of thereof etc.

Design, implementation, and automatic synthesis of numerical feature algorithms for classification are discussed in view of the multipurpose system design. In order to build a specific application, the universal system can be trained for performing specific scene understanding tasks and then applied either in online (real time) or offline situation.

The system can be trained for performing and combining various pattern recognition tasks in real time. Signals of different nature and sources can be recognized by a single system and joined into a simple alerting language. The system is trainable for detecting various objects and situations – on the road, in a production line, in working environment etc.

The proposed system architecture has many application areas and can be an important part for manufacturing specialized devices and systems. The following are typical tasks of existing specialized alerting systems: pedestrian detection, motorcycle detection, bicycle detection, vehicle & lane detection, lane departure warning, forward collision warning, eye fatigue detection. All such applications can be built, with combined functionality, using a single device containing this system.

## Acknowledgement

I am grateful to Dr. V. Shlain for introducing me into the fascinating area of image classification many years ago. Mr. Ilya Kochubeevsky provided some useful links, application ideas and video material. Dr. Ze'ev Ganor made valuable remarks concerning the application issues and the manuscript. Ms. Irena Kemarsky provided valuable help in debugging of the system interface.

## References

1. Vapnik, V. N. The Nature of Statistical Learning Theory (2nd Ed.), Springer Verlag, 2000.
2. Chih-Wei Hsu, Chih-Chung Chang, and Chih-Jen Lin. "A Practical Guide to Support Vector Classification". Department of Computer Science, National Taiwan University, Taipei 106, Taiwan. http://www.csie.ntu.edu.tw/~cjlin
3. R.M. Haralick, K. Shanmugam, O. Dinstain, Textural Features for Image Classification, IEEE TRANSACTIONS ON SYSTEM, MAN AND CYBRNETICS, V.SMC-3, No. 6, 1973, pp. 610-621.
4. Bi-hui Wang; Hang-jun Wang; Heng-nian Qi. "Wood recognition based on grey-level co-occurrence matrix". In: 2010 International Conference on Computer Application and System Modeling (ICCASM), 22-24 Oct. 2010, page(s): V1-269 - V1-272
5. Shuo Chen, Chengdong Wu, Dongyue Chen, Wenjun Tan. "Scene classification based on gray level-gradient co-occurrence matrix in the neighborhood of interest points". In: IEEE International Conference on Intelligent Computing and Intelligent Systems, 2009. ICIS 2009. Issue Date: 20-22 Nov. 2009, pages 482-485.
6. Gleibman, A.H., Intelligent Processing of an Unrestricted Text in First Order String Calculus, M.L. Gavrilova et al. (Eds.): Trans. on Comput. Sci. V, Special Issue on Cognitive Knowledge Representation, LNCS 5540, pp. 99–127, 2009. © Springer-Verlag Berlin Heidelberg 2009, http://dl.acm.org/citation.cfm?id=1573944. Read this paper online at http://www.sampletalk.com/SampletalkLanguage/55400099.pdf.
7. Gleibman, A.H., Knowledge Representation via Verbal Description Generalization: Alternative Programming in Sampletalk Language. In: Workshop on Inference for Textual Question

Answering, July 9, 2005 – Pittsburgh, Pennsylvania, pp. 59–68, AAAI 2005 - the Twentieth National Conference on Artificial Intelligence http://www.hlt.utdallas.edu/workshop2005/papers/WS505GleibmanA.pdf

8. Gleibman, A.H., Reasoning About Equations: Towards Physical Discovery. The Issue of the Institute of Theoretical Astronomy of the Russian Academy of Sciences No.18, 1992, 37 pp. In Russian. http://www.biblus.ru/Default.aspx?book=7b36s38i8

9. Muggleton, Stephen. "Inductive Logic Programming: Issues, Results and the Challenge of Learning Language in Logic". Artificial Intelligence 114: 283–296, 1999.

10. Blass, Andreas; Yuri Gurevich. "Algorithms: A Quest for Absolute Definitions". Bulletin of European Association for Theoretical Computer Science (81), 2003.

11. Robinson J. A. "A Machine-Oriented Logic Based on the Resolution Principle." J. Assoc. Comput. Mach. 12, 23-41, 1965.

12. Peng, H.C., Long, F., and Ding, C., Feature selection based on mutual information: criteria of max-dependency, max-relevance, and min-redundancy, IEEE Transactions on Pattern Analysis and Machine Intelligence, Vol. 27, No. 8, pp. 1226–1238, 2005

13. Hai Nguyen, Katrin Franke, and Slobodan Petrovic, Optimizing a class of feature selection measures, Proceedings of the NIPS 2009 Workshop on Discrete Optimization in Machine Learning: Submodularity, Sparsity&Polyhedra (DISCML), Vancouver, Canada, December 2009.

14. Paul Viola, Michael Jones. "Rapid Object Detection using a Boosted Cascade of Simple Features". TR2004-043 May 2004, IEEE Computer Society Conference on Computer Vision and Pattern Recognition, Copyright © Mitsubishi Electric Research Laboratories, Inc., 2004, 201 Broadway, Cambridge, Massachusetts 02139.

15. Ilya Blayvas, Alfred Bruckstein, Ron Kimmel. "Efficient computation of adaptive threshold surfaces for image binarization". Pattern Recognition 39 (2006) 89 – 101.

16. V. Shlain, A. Gleibman, D. Vagapov. Basic algorithms for an automatic faults detection and classification system in microchip manufacturing control. Probl. Upr. (Problems of Control), 2006, Issue 1, Pages 47–53 (Mi pu308).

17. Smart Cameras: A Review. Yu Shi, Serge Lichman. - Interfaces, Machines and Graphical Environments (IMAGEN), National Information and Communications Technology Australia (NICTA), Australian Technology Park, Bay 15 Locomotive Workshop, Eveleigh, NSW 1430, Australia.

18. Aberbour, M. Mehrez, H. Durbin, F. Haussy, J. Lalande, P. Tissot, A. "A system-on-a-chip for pattern recognition architecture and design methodology". Computer Architectures for Machine Perception, 2000. Proceedings. Fifth IEEE International Workshop, Date: 13-13 Sept. 2000, Page(s): 155 - 162.

19. Jan-Erik Källhammer, Kip Smith, Johan Karlsson, Erik Hollnagel. "SHOULDN'T CARS REACT AS DRIVERS EXPECT?" Proceedings of the Fourth International Driving Symposium on Human Factors in Driver Assessment, Training and Vehicle Design. Autoliv Research, Vårgårda, Sweden; Department of Computer and Information Science, Linköping University, Sweden; http://www.smarteye.se/

20. Smart Car Security System for detecting car thefts (with face recognition): http://www.scribd.com/doc/41706257/Smart-Car-Security-System-Presentation

21. The Mobileye – EyeQ™ vision-system-on a chip: http://www.mobileye.com/en/manufacturer-products/processing-platforms/EyeQ2

22. Fire Detection for Cargo compartment: http://www.mut-group.com/aviation/fire-detection.html

23. Mobile DVR by FSEC: http://electronicsystemsnetwork.com/Documents/MDVR-X%20-%20FSEC.pdf

24. Detecting attention, detecting sleep by Matthew Parks: http://linux01.crystalgraphics.com/view/3183b-NmQ1Y/Detecting_Attention_Detecting_Sleep_flash_ppt_presentation